\documentclass[a4paper, 12pt]{ijamas}
  \usepackage{graphicx}
    \usepackage{algorithm}
    \usepackage[noend]{algpseudocode}
    \usepackage{multirow}
    \usepackage{subcaption}
    \usepackage{epstopdf}
  \usepackage{apacite}
  \begin{document}

  %%%%%%%%%%%%%%%%%%%%%%%%%%%%%%%%%%%%%%%%%%%%%%%%%%%%%%

  \title{Multi Agent Path Finding with Awareness for Spatially Extended Agents}

  \author{\textbf{Shyni Thomas$^1$, Dipti Deodhare$^2$ and M. Narasimha Murty$^3$}}

  \date{$^1$Centre for AI and Robotics(CAIR), DRDO\\
  Bangalore, India\\
  sthomas@cair.drdo.in\\[0.3cm]
  $^2$Centre for AI and Robotics(CAIR), DRDO\\
  Bangalore, India\\
  dipti.deodhare@gmail.com\\[0.3cm]
  $^3$Department of Computer Science and Automation\\
  Indian Institute of Science\\
  Bangalore, India\\
  mnm@csa.iisc.ernet.in}

  \maketitle

  %%%%%%%%%%%%%%%%%%%%%%%%%%%%%%%%%%%%%%%%%%%%%%%%%%%%%%

  \begin{abstract}
  \noindent \emph{Path finding problems involve identification of a plan for conflict free movement of agents over a common road network. Most approaches to this problem handle the agents as point objects, wherein the size of the agent is significantly smaller than the road on which it travels. In this paper, we consider spatially extended agents which have a size comparable to the length of the road on which they travel. An optimal multi agent path finding approach for spatially-extended agents was proposed in the eXtended Conflict Based Search (XCBS) algorithm. As XCBS resolves only a pair of conflicts at a time, it results in deeper search trees in case of cascading or multiple (more than two agent) conflicts at a given location. This issue is addressed in eXtended Conflict Based Search with Awareness (XCBS-A) in which an agent uses awareness of other agents' plans to make its own plan. In this paper, we explore XCBS-A in greater detail, we theoretically prove its completeness and empirically demonstrate its performance with other algorithms in terms of variances in road characteristics, agent characteristics and plan characteristics. We demonstrate the distributive nature of the algorithm by evaluating its performance when distributed over multiple machines. XCBS-A generates a huge search space impacting its efficiency in terms of memory; to address this we propose an approach for memory-efficiency and empirically demonstrate the performance of the algorithm. The nature of XCBS-A is such that it may lead to suboptimal solutions, hence the final contribution of this paper is an enhanced approach, XCBS-Local Awareness (XCBS-LA) which we prove will be optimal and complete.
  }

  \medskip


  \medskip

  \end{abstract}

  %%%%%%%%%%%%%%%%%%%%%%%%%%%%%%%%%%%%%%%%%%%%%%%%%%%%%%%

  \section{INTRODUCTION}\label{introduction}
    \indent The problem of path finding and scheduling of agents over a transportation network attempts to identify paths for the agents to move from their defined start position to their destination position without colliding with any other agent. Collision is defined to occur when two objects occupy the same edge or vertex at the same time. Most of the earlier works have been for point sized agents, which take unit travelling time across nodes of the network. The problem has been attempted using search based approaches using variants of A*~\cite{Standley2011,Silver2005}, Conflict Based Search~\cite{Sharon2015,Felner2018,Boyarski2015}, Propositional logic~\cite{Surynek2019} etc. The problem has also been addressed in a modified form in problems like Automated Guided Vehicle (AGV) routing~\cite{Ling2002} and Vehicle Routing~\cite{Bodin1983}. The variations being defined in terms of constraints imposed on the movement, the characteristics of the moving agents etc.  \\
    
    \indent Approaches for Multi Agent Path Finding (MAPF) while considering the dimensions of the agents have also been attempted where the agents are variously termed as large agents ~\cite{Li2019}, multi-sized agents ~\cite{Harabor2008}, train agents ~\cite{Atzmon2019} and spatially extended agents~\cite{shivashankar2009}. Real world versions of the problem includes Convoy Movement Problem~\cite{Chardaire99,Ramkumar2015,Khemani2012,Thomas2008}, Train Scheduling~\cite{Bettinelli2017,Wang2018} etc. MAPF for spatially extended agent is more challenging than point objects as (\textit{i}) At a given time instance, spatially extended agents may occupy multiple edges of a road network, (\textit{ii}) Edge transitions are durative and equals the time taken by the complete length of the agent to cross over, (\textit{iii}) Conflicts can occur on edges when two agents cross over in opposite directions, or one overtakes another in the same direction. Conflicts can also occur on vertices when two agents cross over.\\
    
    \indent In a new approach to the problem, eXtended Conflict Based Search with Awareness (XCBS-A) \cite{Thomas2018} was proposed to handle multiple conflicts simultaneously thereby leading to faster convergence to the solution. To identify sets of mutually conflicting agents, the paper proposes a new spatio-temporal data structure namely, Temporal Occupancy (T.O.) Graph. The conflict resolution strategy is based on awareness of other agents plans. In this approach, the search space is maintained as a Constraint Tree (CT); exploration of any node of the CT leads to a maximum of $i^j$ child nodes being generated, where $j$ is the number of conflicting blocks and $i$ being the number of agents per block, resulting into a huge search space.  \\
        
    \indent The first contribution of this paper is a theoretical evaluation of XCBS-A and experimental evaluation of its performance in terms of variances in agent, road and plan characteristics. To prove the claim of a distributed algorithm we evaluated the algorithm by distributing the agents on multiple machines and compared performance. Our next contribution is a methodology for reducing the search space, where we only add selective CT-nodes to the Open Set to be explored and retain the remaining in a Potential Set; three heuristic functions are defined to extract the nodes from Potential Set into the Open Set and the approach is empirically evaluated for efficacy.\\
    
    \indent As our third contribution in this paper, we address the second concern of XCBS-A namely optimality. XCBS-A results in suboptimal solutions primarily because it assumes plans of other agents in a block to be fixed, down the branch of the CT. We propose a revision to this approach and call it as XCBS-LA i.e. XCBS with Local Awareness. We prove that the algorithm guarantees optimality and completeness. \\
    
    \indent This paper is organised to include a brief overview of related work in Section \ref{related_work}, we then define the terms used in the paper in Section \ref{terminology}. Sections \ref{xcbsa}, \ref{xcbsa-theoretical} \& \ref{xcbsa-experimental} cover a description of the XCBS-A algorithm, theoretical analysis and empirical evaluation. In Section \ref{xcbsa-eff}, we propose the heuristics for reducing the search space and bring out the XCBS-LA algorithm in Section \ref{xcbsla}. The conclusions drawn from the effort have been covered in Section \ref{conclusion} of the paper.
    
  %%%%%%%%%%%%%%%%%%%%%%%%%%%%%%%%%%%%%%%%%%%%%%%%%%%%%%%

    \section{RELATED WORK}\label{related_work}
    \subsection {Multi Agent Path Finding (MAPF)} The MAPF problem has been handled in various varieties for point based objects. One of the notable work was the approach of Conflict Based Search~\cite{Sharon2015} which proposed an optimal, complete path planning solution for multiple agents using a two level search on a Constraint Tree. A low level planner would find consistent plans for individual agents satisfying constraints imposed on each of them individually.  A high level planner then would search the Constraint Tree for valid (non-conflicting) solutions. There have been several variations and extensions to the approach. Some of the notable ones include meta-agent based CBS~\cite{Sharon2012} in which agents with multiple conflicts are grouped together. Another approach is the Improved Conflict Based Search (ICBS)~\cite{Boyarski2015} in which plans were restarted for merged agents and conflicts were prioritised as cardinal, semi-cardinal and non-cardinal cases and handled accordingly. Another complete algorithm for multi-agent planning was the Push and Rotate approach~\cite{De2014} where conflicting agents on a biconnected graph were de-conflicted for plan generation by executing push and rotate actions, leveraging the concept of bridge on the graphs.\\
    
    \indent There have been some attempts on planning for agents of non-point size. Planning for agents with definite geometric shapes, occupying multiple locations was attempted in the algorithm Multi-Constraint Conflict Based Search (MC-CBS)~\cite{Li2019}. In another approach~\cite{Atzmon2019}, agents which occupied multiple locations at the same time instance owing to their length, termed as train agents, were planned for. The agent transitions from one location to another were instantaneous in nature. The approach handled two types of conflicts i.e. self conflict when an agent conflicts with itself, and occupation conflict, when an agent occupied a location, which was occupied by another agent at the given time instance. The solution proposed, Multi-Train Conflict Based Search (MT-CBS) was developed along the lines of CBS, the variation being at the low level planning and conflict detection approach. A continuous time formulation of the problem was proposed in Continuous-Time Conflict Based Search (CCBS)~\cite{Roni2019}, an adaptation of CBS and a customized version of Safe Interval Path Planning (SIPP), to arrive at a sound, complete and optimal solution to the problem.\\
    
    \indent Spatially extended agents also find manifestation in real-world problems like train scheduling and Convoy Movement Problem (CMP). A brief overview on the centralised as well as distributed approaches adopted for CMP is covered in the following paragraphs.

    \subsection{Convoy Movement Problem} The CMP was proposed and solved using an integer programming (IP) formulation based on the concept of a time-space network with a combination of Lagrangian Relaxation and heuristic-based optimization techniques~\cite{Chardaire99,Harrison2000}. Kumar et.al. arrived at the lower bounds of the CMP in ~\cite{RamKumar2011}. Several other integer programming formulations have been formulated for the approach including ~\cite{RamKumar2008} and~\cite{Bovet91}.\\
    
    \indent While formulating CMP for peacetime operations with the objective of minimizing total travel time and the travel space, goal programming and simulated annealing based approaches was explore in ~\cite{RamKumar2009}. The solutions were demonstrated on small and medium sized problems. Another look into the peacetime movement of convoys was attempted where the objective was to minimize civilian traffic disruption~\cite{Sadeghnejad2017}. The paper proposed an exact hybrid algorithm that combines the \textit{k}-shortest path algorithm, finding a minimum weighted \textit{k}-clique in a \textit{k}-partite graph and branch-and bound strategy. In~\cite{Robinson2006} a genetic algorithm was adopted with a discrete event simulation and Swarm algorithm based approaches have been looked into in~\cite{Ramkumar2015,Lee96}.\\
    
    \indent Distributed approaches to the problem were attempted, in~\cite{Thangarajoo2010}, using a hybrid conflict resolution by coalition formation and distributed constraint optimisation approach. The concept of reservations on the time-windows of the network resources for convoy agents was used in~\cite{Adriaan2009}. This approach assumed point nature of the moving object. An A* based approach was discussed in~\cite{Thomas2008}. A centralised plan was generated by a state space search on the states achieved by the convoys on occurrence on each transition event. The approach demonstrated achievement of optimal results. A Planning Domain Definition Language (PDDL) based planning approach was explored in~\cite{Khemani2012} wherein the problem was first captured as a PDDL domain and then using standard planners, they demonstrated results for different problem sets.\\
    
    \indent Multi agent based approaches have also been attempted for the problem. In~\cite{Thomas2015}, eXtended Conflict Based Search (XCBS) was proposed with two levels of search. The first level is at a local plan level by individual agents. The second level of search for optimised solution was at the level of a centralised agent which attempts to optimise the overall solution. 

    \subsection{Multi Agent Conflict Detection}
    For the planning of the spatially extended agents, given that conflict detection is one of the prime concerns, a brief look into the various approaches which have been adopted for conflict detection in literature is in order. In~\cite{Silver2005} a concept of reservation table was introduced where following a predefined ordering an agent would plan its path and ensure it did not conflict with the plans stored by the agents before it in the reservation table. In~\cite{Standley2011}, a conflict avoidance table is used to show the positions of all agents at a given time instance. For point objects, such approaches will enable conflict detection by linear search over all the time-instances. In XCBS~\cite{Thomas2015},  the conflict detection was achieved by dividing the entire plan-time into equal sized time intervals. The spatial location of every pair of agents was compared to check for conflict. In MT-CBS,~\cite{Atzmon2019}, the conflict detection for train agents is accomplished by defining an occupation-list, a list of locations occupied by an agent at an instance of time. As transition from one location to another is instantaneous, the occupation list is checked for every pair of agents at every instance of time. Overlapping occupation list indicated a conflict.

  %%%%%%%%%%%%%%%%%%%%%%%%%%%%%%%%%%%%%%%%%%%%%%%%%%%%%%%

    \section{Terminology}\label{terminology}

    Table \ref{tab:notations} refers to the notations used in this paper.
    \begin{table}[h]
    \resizebox{\textwidth}{0.35\textwidth}{
	    \begin{tabular}[c]{|l|p{3.5cm}|p{8.5cm}| }
		    \hline
		    \multicolumn{1}{|c|}{\textbf{ }}&\multicolumn{1}{|c|}{\textbf{Notation}}&\multicolumn {1}{|c|}{\textbf{Description}} \\
		    \hline
		    \multirow{3}{*}{Network}&$\mathcal{G(V,E)}$ &Road network with $\mathcal{V}$ vertices and $\mathcal{E}$ edges \\
		    &$\mathcal{L}_i$ &length of edge $e_i$, $e_i \in \mathcal{E}$ \\
		    &$\mathcal{S}_i$ &speed on edge $e_i$, $e_i \in \mathcal{E}$ \\
		    \hline
		    \multirow{6}{*}{SEAgent}&$A$ &Set of spatially extended agents \\
		    &$a_j$ &Spatially Extended Agent \\
		    &$\mathbb{L}_j$ &Length of the agent \\
		    &$\mathbb{I}_j$ &Initial (Source) vertex\\
		    &$\mathbb{F}_j$ &Final (Destination) vertex \\
		    &$\mathbb{S}_j$ &Average Speed \\
		    &$\mathbb{E}(t)$ = $\lbrace l_1, l_2, ....,l_n \rbrace$, $l_i \in (\mathcal{E} \cup \mathcal{V})$ &locations occupied by $a_j$ \\ 
		    \hline
		    \multirow{2}{*}{Agent Actions}&move($l_i$, $t$) &head($a_j$) enters location $l_i$ at time $t$ \\
		    &wait($t$, $t_d$) &$a_j$ waits at time $t$ for $t_d$ time-units with no change in the position.\\
		    \hline
		    \multirow{2}{*}{Plan}&$\mathtt{P_\textit{j}} = \{\mathit{move(l_1,t_1)}$ , $\mathit{move(l_2, t_2)}$, $\mathit{wait(t_3, d) \ldots\}}$, 
		    
		    where, $t_1<t_2<t_3$&The sequence of actions to be executed by an agent $a_j$ to travel from $\mathbb{I}_j$ to $\mathbb{F}_j$. A plan $P_j$ is \textit{consistent}
		    if it satisfies all constraints imposed on $a_j$. \\
		    \hline
		    \multirow{2}{*}{Solution}&$\mathtt{S} = \{\mathtt{P}_1, \ldots ,\mathtt{P}_k\}$ & Solution $S$ should satisfy: \textit{(i)} $\forall{j_{1..k}}\mathtt{P}_j$ is consistent and \textit{(ii)} $\mathtt{S}$ is \textit{valid} i.e. plans are conflict-free.\\
		    \hline
	    \end{tabular}
	    }
	    \caption{Description of notations used in this paper}
	    \label{tab:notations}
    \end{table}
  %%%%%%%%%%%%%%%%%%%%%%%%%%%%%%%%%%%%%%%%%%%%%%%%%%%%%%%

    \section{XCBS-A}\label{xcbsa}
    \indent We now describe the eXtended Conflict Based Search with Awareness (XCBS-A), which is based on the principle of the two-level search used in XCBS. The algorithm proposes a novel approach for conflict detection, low level search and conflict resolution. 
    \subsection{Definitions}
    The definitions used in this approach are as follows:
    \begin{definition}
    The search space for the plans is maintained in a \textbf{Constraint Tree} (CT). A node of the CT, the CT-node $N$, is consistent with the definition given in XCBS~\cite{Thomas2015}. It comprises: 
    \begin{enumerate}
	    \item \textit{N.Id}: Unique identifier for the CT-node.
	    \item \textit{N.Solution}: $\{\mathtt{P_\textit{1}},\ldots ,\mathtt{P_\textit{k}}\}$ is a set of consistent optimal plans $\mathtt{P_\textit{i}}$ for each agent. 
	    \item \textit{N.Cost}: Cumulative cost of the plans of the individual agents i.e.,$ \sum_{j=1}^k cost(\mathtt{P}_j)$ 
    \end{enumerate}
    \end{definition}

    \begin{definition} The coordination for the search to achieve a valid solution is done by the \textbf{Central Agent} (CA). The CA executes a high-level search on the Constraint Tree (CT) to identify potential nodes to be explored for validity of solution.
    \end{definition}

    \begin{definition} Every agent for which the path is to be planned is termed as the \textbf{Spatially Extended Agent} (SEA). The agent executes a low-level search to generate a consistent plan with respect to the spatio-temporal constraints imposed by its own preferences as well as other agents' plans.
    \end{definition}

    \begin{definition}The \textbf{Temporal Occupancy (t.o.) node} $\gamma$, expressed as a tuple $<l,\tau, Z>$, is used to describe the occupancy details of a location $l$ during a given time period $\tau$. The list of exact time durations, during $\tau$, during which each SEA occupies the location $l$ is given by AgentDetails $Z$. There exists at least one t.o. node, for each location $l \in \mathcal{E} \cup \mathcal{V}$, which is in any plan $\mathtt{P}_j \in \mathtt{S}$. For a given location $l$, if the time-intervals spent by different agents on $l$ overlap, then $\tau_{l}$ for the corresponding t.o. node $\gamma_{l}$ will be defined as the maximal covering time interval, else multiple t.o. nodes will be generated for the edge or vertex $l$.
    \end{definition}

    \begin{definition} A directed graph $G(\Gamma,\Delta)$ is called the \textbf{Temporal Occupancy (T.O.) graph} comprising $\Gamma$, the set of t.o. nodes, and $\Delta$, the set of edges which connects the t.o. nodes. A directed edge $\delta \in \Delta$, exists between two t.o. nodes, $\gamma_i$ and $\gamma_j$, when $\gamma_i.g$ and $\gamma_j.g$ are distinct and are consecutive locations in the plan $\mathtt{P_j}$ of agent $a_j$. Each edge $\delta$ has a property $\delta.agent$ to indicate the agent whose plan makes that edge valid. Two t.o. nodes may have multiple edges between them, if more than one agent occupies the corresponding $\gamma_i.g$ during the time periods of the t.o. nodes.
    \end{definition}
    The T.O. graph may contain multiple t.o. nodes for which $\gamma_i.g$ is same. For such $\gamma_i$, $\gamma_i.\tau$ will be governed by the meets, meets inverse, after or before relation as defined in Allens temporal relations~\cite{Allen1983}. These nodes indicate occupancy details over a road over non-overlapping time-intervals. At its initial and final vertices $\mathbb{I}_j$, $\mathbb{F}_j$, an agent cannot collide with any other agent. For this, TO nodes are not generated for the agent for its initial and final vertices. The graph can also have several singleton t.o. nodes, which can occur when the path of the agent comprises a single edge between its initial and final vertices.    

    \subsection{Algorithm}

    \subsubsection{High Level Search} 
    The High Level Search, shown in Algorithm \ref{xcbsa-highlevelsearch}, is executed by the CA, on the search space of the plans of all the agents, to arrive  at a valid solution. The CT, which maintains the search space of the plans, is initialised with the root CT-node, with $N_r.Solution$ having the initial plans $\mathtt{P_\textit{i}}$ of SEAs, and cost $N_r.Cost$. The initial plans are made assuming that the entire road network is fully available to each agent. The Open Set $O$, comprising the unexplored CT-nodes is populated with the CT-nodes generated (initially it will have only the root CT-node, as in Line 2). At Line 5, the CA chooses the least cost CT-node, $N_e$, for exploration from $O$. This CT-node, $N_e$, is validated to identify presence of conflicts between the plans of the agents, as discussed in Algorithm \ref{Validate CT-node}. If conflicts are detected, the conflicting plans are sent to the SEAs for resolution. The resolution plans generated by the SEAs through a Low-Level search are combined into child CT-nodes, discussed under section \ref{conflictResolution}, which are added to $O$ for further exploration. The search is continued till a conflict-free CT-node is found. \\

    \begin{algorithm}
	    \caption{\textbf{High Level Search}}\label{xcbsa-highlevelsearch}
	    \begin{algorithmic}[1]
		    \State Define root CT-Node $N_r$ with optimal consistent plan for each $a_i$ agents.
		    \State OpenSet $O \gets \{N_r\}$ 
		    \State $ResultFound \gets false$.  
		    \While {!$O.empty$ or !$ResultFound$} 
		    \State $N_e \gets arg min_{N^{'} \in O} N^{'}.Cost$
		    \State $O \gets O \setminus {N^{'}}$
		    \State $conflictPresent \gets Validate(N_e)$
		    \If {$conflictPresent$}
		    \State $\{ChildCTNodes\} \gets ResolveConflict(N_e)$
		    \State $O \gets O \cup \{ChildCTNodes\}$
		    \Else 
		    \State $ResultFound \gets true$
		    \EndIf
		    \EndWhile  
		    \If {$ResultFound$}
		    \State $Solution \gets N_e.Solution$
		    \EndIf
	    \end{algorithmic}
    \end{algorithm}

    \subsubsection{Solution Validation}\indent Solution validation checks for presence of conflicts between the plans in the CT-node $N_e$ being explored and identifies all mutually conflicting sets of agents. A solution is valid if no conflicts are detected among the plans in $N_e$. Solution validation involves three stages: \textit{(a)} Creation of TO Graph $G_e$, \textit{(b)} Creation of partition set, \textit{(c)} Decision on validity of solution. We step through each stages in the following paragraphs.\\
    
    \indent Stage (a): For the CT-node $N_e$ being explored, a TO graph $G_e$ is created as shown in Lines 1-18 of Algorithm \ref{Validate CT-node}. The t.o.node, $\gamma$, represents the occupancy information of every location for periods of time. Every location in the plan of an agent $a_i$, is compared with the previous t.o. nodes to check if a t.o. node, $\gamma$ exists for the given location. If so, $\gamma.Z$ and  $\gamma.\tau$ is updated to include the new agent and covering time-period. In case such a t.o. node does not exist, then creation of a t.o. node with the relevant values is done. Further, every t.o. node so created/updated, is then linked to the previous t.o. node through an edge $\delta$ with the relevant agent property being set. In the TO graph so created, t.o. nodes with multiple incoming and outgoing edges indicate that there are multiple agents occupying that particular edge at the given time period and hence possibility of collision needs to be evaluated only for those nodes.\\
    
    \indent Stage (b): Lines 19-26 of the Algorithm \ref{Validate CT-node} cover the creation of partition $X$,  from $G_e$. From the list of the t.o. nodes with in/out degree $>$ 1, the overlapping agents are added to the same block of the partition. When adding to a block, if none of the agents pre-existed in any block then a new block is created with both agents as members. If either one of the agents is a member of some pre-existing block of the partition, then the other agent is also added to the same block. If both agents pre-exist and belong to different blocks, then the two blocks are merged into a single block. Once all the t.o. nodes have been evaluated, if there are still agents which do not belong to any block of the partition, they are created into singleton blocks.\\
    
    \indent Stage (c): The non-singleton blocks of the partition comprise the mutually conflicting agents. The $N_e.Solution$ is valid when the plans are conflict free. When a partition $X$ comprises only of singleton blocks, it indicates that the plans are consistent and conflict free and hence the solution of the CT-node is valid.\\
    
    \begin{algorithm}
    	    \caption{\textbf{Validate $N_e$}}\label{Validate CT-node}
    	      \begin{algorithmic}[1]
		    \Procedure{Validate}{$N_e$}
		    \For {plan $p_i \in N_e.Solution$}  \Comment {Stage (a): Create TO Graph $G$}
		    \State $prevNode \gets null$,  $currNode \gets null$
		    \For {location $l_{ij} \in p_i$}
		    \If {$\exists \gamma \in G.\Gamma, (\gamma.g = l_{ij})\bigwedge (l_{ij}.TimeDuration$ overlaps $\gamma.\tau) $}
		    \State $\gamma.Z.agent \gets agent_i  \cup \gamma.Z.agent$
		    \State $\gamma.Z.AgentTime \gets l_{ij}.TimeDuration \cup \gamma.Z.AgentTime$
		    \State $\gamma.\tau \gets covering(\gamma.\tau, l_{ij}.TimeDuration)$
		    \Else    
		    \State Create t.o.node $\gamma$
		    \State Add $agent_i$ to $\gamma.Z.agent$
		    \State Add $l_{ij}.TimeDuration$ to $\gamma.Z.AgentTime$
		    \State $\gamma.\tau \gets l_{ij}.TimeDuration$
		    \State Add $\gamma$ to $G$
		    \EndIf
		    \State $currNode \gets \gamma$
		    \If {$prevNode != null$}
		    \State Create edge $\delta \in G.\Delta$, between $prevNode$ and $currNode$
		    \State $\delta.agent \gets agent_i$ ,  $prevNode \gets currNode$, $currNode \gets null$
		    \EndIf
		    \EndFor
		    \EndFor
		    
		    \State Create blank partition $X$. \Comment Stage (b): Create Partition on $G$
		    \For {t.o. node $n_e \in G, degree(n_e) > 1$}
		    \For {$a_i, a_j \in n_e.Z.agent$}
		    \If {$Relates(n_e.Z_{a_i}.AgentTime, n_e.Z_{a_j}.AgentTime)$}
		    \If {$\exists b \in X, a_i \lor a_j \in b$} 
		    \State Update $b$ to include $a_i$, $a_j$.
		    \Else 
		    \State Create block $b$ ,  Add $a_i$, $a_j$ to $b$.
		    \EndIf
		    \EndIf
		    \EndFor
		    \EndFor
		    
		    \If {$\exists a_i \in \mathbb{A}, \forall b \in X, a_i \not\in b}$  \Comment Add singleton blocks to the partition.
		    \State Add singleton block $b$ containing $a_i$
		    \EndIf
		    
		    \If {number of non-singleton blocks in $X \geq 1$} \Comment Stage (c): Decision.
		    \State return true.
		    \Else
		    \State return false.
		    \EndIf
		    \EndProcedure
	    \end{algorithmic}
 
    \end{algorithm}

    \indent  In XCBS, in case of $t$ time  intervals for $k$ agents, an agent could occupy $l$ locations during that interval; the time taken for detection would be $O(t\cdot k^2 \cdot l)$. However, in our approach of using TO graph for conflict detection, we need to consider the time taken for creation of the TO graph as well as the time for consequent conflict-detection.  The creation of the TO graph requires iteration over the plan of each of the $k$ agents. For each agent, if there are $l$ locations in the plan, then the creation time is $O(k\cdot l)$. Consequent to the creation of the TO graph $G$, only the set of t.o. nodes with an in-degree or out-degree $>$ 1 or $AgentDetails > 1$ need to be evaluated for checking the conflict time. If there are $n$ such conflicting t.o. nodes, the overall time for conflict-detection (including TO graph creation time) is $O(k \cdot l + n \cdot k^2)$.

    \subsubsection{Conflict Resolution}\label{conflictResolution}

    \begin{algorithm}
	    \caption{\textbf{ResolveConflict $N_e$}}\label{resolveConflict}
	    \begin{algorithmic}[1]
		    \Procedure{ResolveConflict}{$N_e$}
		    \For {each agent $a_i$ in non-singleton blocks} 
		    \State $P^{'}_{i}$ = LowLevelSearch($a_i, N_e$).
		    \EndFor 
		    \For {each combination on the members of non-singleton block $b$ in the partition of $N_e$} 
		    \State Create child CT-node $N_c$ with revised plan of one agent $a_i$ from each of $b$ and the unchanged plans for the remaining agents.
		    \State Compute Cost($N_c$).
		    \State Add to the openset $O$
		    \EndFor 
		    \EndProcedure  
	    \end{algorithmic}
    \end{algorithm}

    \indent When a conflict is detected on a CT-node $N_e$, each SEA in a non-singleton partition block invokes a LowLevelSearch for itself to generate an alternate plan, as shown in line 2 of Algorithm \ref{resolveConflict}. All agents' plans are passed to each of the SEA to generate the alternate plan such that it does not conflict with the other agents' plans. The LowLevelSearch will always be able to generate an alternative plan either by deviation in terms of path or by introducing a wait before the conflict location. When revised plans $P^{'}_{i}$  are received from the SEAs, the Central Agent composes the revised plans into child CT-node, as shown in Line 4. If $N_e$ resulted in $j$ non-singleton blocks of partition, one agent's revised plan is added for each non-singleton block and the plans of the remaining agents of the same block remain unchanged in a generated child CT-node.  The plans for the singleton block agents are added unchanged. Thus, if $N_e$ resulted in $j$ non-singleton blocks of partition then, each child CT-node will have $j$ revised 
    plans in $N.Solution$. A combination on the agents in the non-singleton blocks is done to generate child CT-nodes. Thus, for $j$ non-singleton blocks with $i$ agents each, the number of CT-nodes generated will be $i^j$. \\
    
    \indent For the revised plan generated by an SEA in each of the child CT-node generated by the CA, a conflict can only potentially arise between the new $j$ revised plans generated independently by the agents in $j$ blocks of the partition. The child CT-nodes so generated, are added to the Open-Set $O$ to be further explored. \\

    \subsubsection{Low Level Search}
    \begin{algorithm}
	    \caption{\textbf{Low Level Search $N_e$}}\label{LowLevelSearch}
	    \begin{algorithmic}[1]
		    \Procedure{Low Level Search}{$N_e$}
		    \If {!agentPlansAvailable}
		    \State Get shortest path from $\mathbb{I}_i$ to $\mathbb{F}_i$
		    \State Define plan $P_i$ with the above path.
		    \Else 
		    \State Get constraints set from the plans of other agents.
		    \State $R_r \gets$ CreateRootReplanNode().
		    \State OpenSet $O \gets R_r$ ;  $O \gets O \setminus \{R_{r}\}$;$ResultFound \gets false$.  
		    \While {!$O.empty$ or !$ResultFound$} 
		    \State $R_e \gets arg min_{R^{'} \in O} R.cost $
		    \State $consistent \gets checkConsistency(R_e)$
		    \If {$!consistent$}
		    \State $Child_1 \gets createAltPath(R_e)$ ;  $Child_2 \gets createWait(R_e)$
		    \State $O \gets O \cup \{Child_1, Child_2\}$
		    \Else 
		    \State $ResultFound \gets true$
		    \EndIf
		    \EndWhile  
		    \EndIf
		    
		    \EndProcedure
	    \end{algorithmic}
    \end{algorithm}

    \indent Each SEA executes a Low-Level search to identify consistent plans for itself, as shown in Algorithm~\ref{LowLevelSearch}. During the first run of the algorithm (no other agent plans are available), each SEA generates its optimal path using Djikstra's algorithm~\cite{Djiktstra1959} assuming full availability of road-space, as in Lines 2-4. The plan so obtained is the best plan for each agent. In subsequent iterations of the algorithm (Lines 6-16), each SEA identifies spatio-temporal constraints which are imposed on it by the current known plans of the other agents. The SEA then generates an alternate plan $P^{'}$, to meet all the newly generated spatio-temporal constraints. Consistency with respect to plans of other agents is maintained either by identifying an alternate location or by introducing a wait. The optimal plan with no conflict is sent to the CA as the alternative plan. The new plan so generated is different from the initial plan either spatially or temporally or both. This change leads to the cost of the new plan being same or higher than its earlier plan. \\
    \begin{figure} 
    \centering
    \begin{subfigure}{0.6\textwidth}
	    \includegraphics[width=1.2\textwidth]{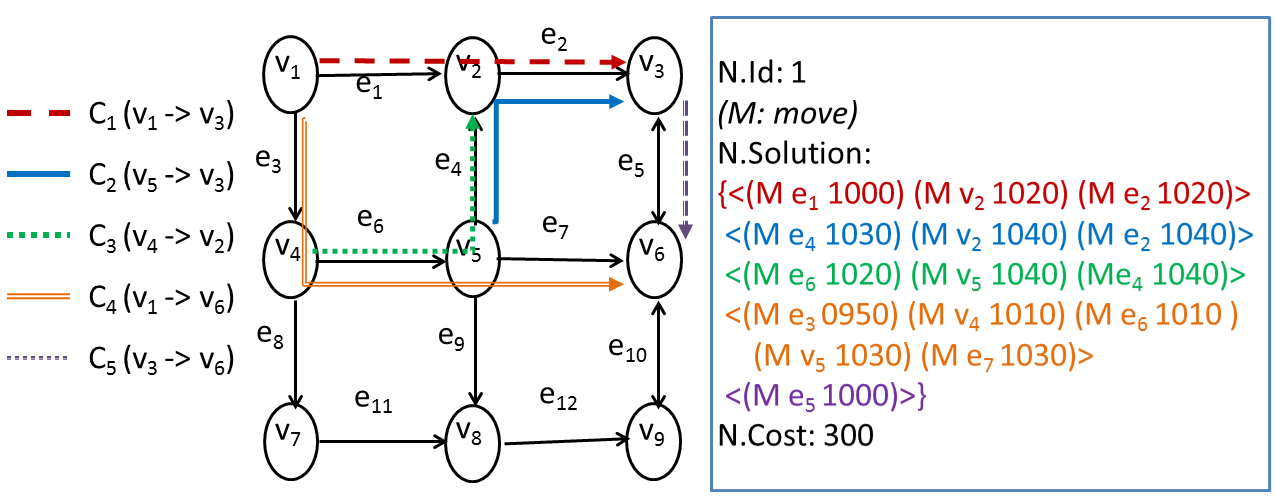}
	    \centering
	    \caption{Sample network with root CT node $N_r$}
	    \label{fig:roadgraph}
    \end{subfigure}
    \begin{subfigure}{.5\textwidth}
	    \includegraphics[width=0.9\textwidth]{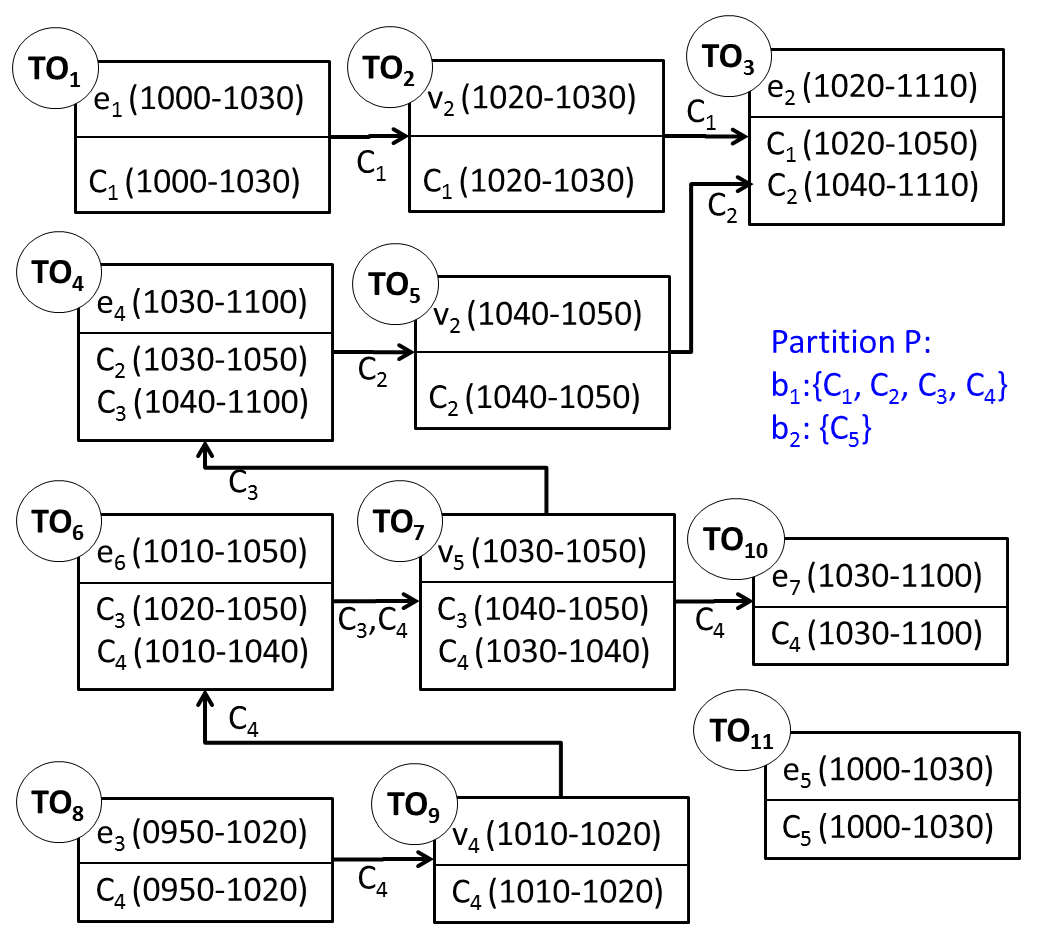}
	    \centering
	    \caption{TO Graph with partition of $N_r$ }
	    \label{fig:TOGraph}
    \end{subfigure}%
    \begin{subfigure}{.5\textwidth}
	    \centering
	    \includegraphics[width=0.9\textwidth]{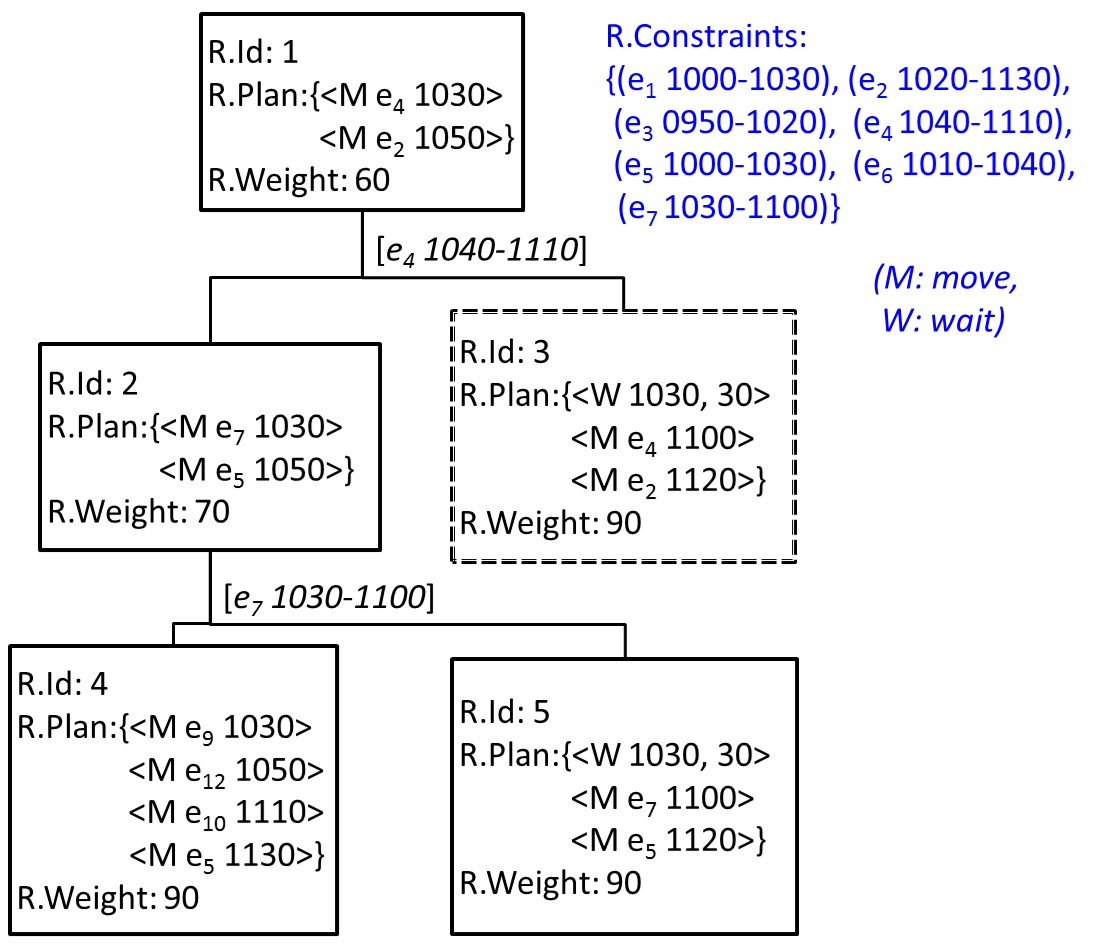}
	    \caption{Replanning for agent $C_2$}
	    \label{fig:Replan}
    \end{subfigure}
	    \caption{Figure shows the TO Graph and the replanning at an agent level, as a consequence of validation of $N_r$ }
	    \label{fig:Replan_TOGraph}
    \end{figure}

    \subsection{Example} We will walk through an example scenario with five SEAs moving in a grid network, as shown in Figure \ref{fig:roadgraph}. There is a t.o. node created for every unique location which is a part of the plans of any SEA. For SEA $C_1$, travelling through $e_1,v_2,e_2$, the t.o. nodes $TO_1, TO_2$ and $TO_3$ are created. When $C_2$ is evaluated, it passes through $e_4, v_2$ and $e_2$ and hence t.o. node $TO_3$ is updated to include $C_2$ in its agents' list and the time duration is the covering time-period of the two agents. Similarly $C_5$ which travels through only one edge $e_5$ creates a singleton t.o. node $TO_{11}$. Using the TO graph, the partition $X$ of the agents is created. $C_1$ shares location $e_2$ with $C_2$, similarly, $C_2$ and $C_3$ conflict on edge $e_4$, and $C_3$, $C_4$ collide on edge $e_6$, thus any change in the schedules of $C_3$ may have cascading impact on $C_1, C_2$ or $C_4$, hence they are put in the same block $b_1$ of the partition. As $C_5$ does not intersect with any other SEA, it is put into singleton block $b_2$ of the partition $X$.  Thus, for the CT-node, a partition with two blocks is generated; the TO Graph and the partition is shown in Figure \ref{fig:TOGraph}. The block $b_1$, has agent $C_2$ as a member, and the tree for replanning, generated as part of the Low-Level search for agent $C_2$ is shown in Figure \ref{fig:Replan}. The spatio-temporal constraints on $C_2$ from the plans of the other agents is shown as R.Constraints. The root Replan-node (R.Id:1), contains the current plan for $C_2$. From the constraint set, the first identified constraint that gets violated is road $e_2$. Replan-nodes 2,3 handle the repath and the waiting plans respectively. Replan node 2 being the lower cost node is explored next and found to violate constraint with respect to road $e_7$ and hence resolved to generate Replan-nodes 4 and 5. The final plan which is consistent with respect to all other agent plans is shown by the highlighted bordered node (Id:3)\\

    %%%%%%%%%%%%%%%%%%%%%%%%%%%%%%%%%%%%%%%%%%%%%%%%%%%%%%%

    \section{ANALYSIS}\label{xcbsa-theoretical}

    \begin{lemma}
    The number of singleton blocks in the partition of a CT-node is non-decreasing down the branch of a CT.\\
    \end{lemma}
    \indent Let $N_0$ be the node at the head of a branch of the CT, with its partition set containing $s$ singleton blocks and $c$ non-singleton blocks. In the child CT-node $N'_0$, the singleton nodes remain unaffected; however, one element from each of the $c$ blocks will be reduced because of conflict resolution. These $c$ elements will either conflict with each other, or form singleton blocks. Thus the number of singleton blocks in the child CT-node will be either $s$ i.e. same as the parent CT-node or will increase by a maximum of $c$. \\

    \begin{lemma}
    The average size of non-singleton blocks is non-increasing at every level down a branch in the CT. \\
    \end{lemma}

    \indent Let the total number of non-singleton blocks in a partition of the root CT-node be $c$, with an average size \textbar${c}$\textbar, and the number of singleton blocks be $s$. If the number of agents be $k$ then\\
    \begin{equation*}
    |c| *c + s = k 
    \end{equation*}
    \begin{equation}
    |c| = (k - s)/c
    \end{equation}

    After the first step of resolution, one element from each of the non-singleton blocks is revised and removed from the block, reducing the size of the non-singleton blocks by one. Each of these plans can either all conflict among themselves to generate a new additional block in the partition, or can have no conflicts among themselves leading to independent singleton blocks, or can form multiple additional blocks. We will consider the three boundary cases:

    \textit{Case 1:} No new conflicting blocks due to revised plans are introduced in the child CT-node partition.\\
    \indent In this case, consequent to revised plans, one element from each of the $c$ blocks will form a singleton block by themselves.\\
    \begin{equation}
    \implies s' = s + c      
    \end{equation}

    since $|c'|*c' + s' = k$ \\
    \begin{equation*}
    \implies |c'| = (k- s')/ c' \\
    \end{equation*}

    from (3), \begin{equation}|c'| = (k - s -c)/c'
    \end{equation}

    As the number of non-singleton blocks are unchanged from the previous step, hence $c' = c$ \\
    $\implies |c'| = (k - s -c)/c \\$
    $\implies |c'| = (k - s)/c - 1 \\$
    $\implies |c'| = |c| - 1 \\$

    \textit{Case 2:} The revised plans all mutually conflict to introduce one additional non-singleton block in the child CT-node partition.\\
    In this case, consequent to revised plans, number of non-singleton blocks $c'$ will be $c' = c+ 1$\\ 
    As the number of singleton blocks are unchanged from the previous step, hence $s' = s$ \\
    $\implies |c'| = (k - s')/c'$\\
    $\implies |c'| = (k - s)/(c + 1)$\\
    $\implies |c'| < |c|$\\

    \textit{Case 3:} The revised plans all pairwise conflict to introduce $c/2$ additional non-singleton block in the child CT-node partition.\\
    In this case, consequent to revised plans, number of non-singleton blocks $c'$ will be $c' = c+ c/2$\\ 
    As the number of singleton blocks are unchanged from the previous step, hence $s' = s$ \\
    $\implies |c'| = (k - s')/c'$\\
    $\implies |c'| = (k - s)/(3/2 c)$\\
    $\implies |c'| < |c|$\\
    \begin{lemma}
    The Cost of a CT-node is non-decreasing down a branch of the CT.
    \end{lemma}

    \indent The evaluated CT-node $N_e$ has $N_e.Solution = \{P_1,\ldots, P_k\}$ where,\\

    $\forall_{i=1 \ldots k}P_i = \min( cost(P_{i_1}),\ldots,cost(P_{i_j}))$.\\
    $\implies \nexists \bar{P}_i \colon cost(\bar{P}_i) < cost(P_i)$.\\

    In other words, $N_e.Solution$ comprises the least cost plans of the individual agents.\\
    The child CT-node $\acute{N}_e$ of $N_e$ may have $m$ agents with revised plans $\acute{P}_i$ generated to avoid conflict with other agents' plans. Further, $cost(\acute{P}_i) >= cost(P_i)$, as the revised plan for the agent will be generated to handle the additional spatio-temporal constraints imposed due to potential conflict.\\
    $\implies cost(\acute{N}_e.Solution) >=cost(N_e.Solution)$. \\

    \begin{theorem} XCBS-A is complete. \end{theorem}
    Proof: The CT contains all the possible consistent plans of the agents as CT-nodes, which are explored based on their cost. The algorithm systematically explores all equal cost CT-nodes, before choosing a higher cost CT-node, for possible valid solution. Hence it is akin to a uniform-cost search and hence is complete. Further, the algorithm keeps resolving the partition set of the root CT-node to arrive at singleton sets, down a branch of the CT. From Lemmas 1\&2, we know that the partition eventually resolves to singleton sets and hence cannot be resolved further. Thus each branch of the CT will lead to a leaf CT-node and hence the algorithm will terminate either when a solution is found or when there are no-more CT-nodes to be explored. 
  %%%%%%%%%%%%%%%%%%%%%%%%%%%%%%%%%%%%%%%%%%%%%%%%%%%%%%%

    \section{EXPERIMENTAL ANALYSIS}\label{xcbsa-experimental}
    We have evaluated the time taken to arrive at a valid solution and the search space explored by XCBS-A with respect to some of the other available solutions. The road networks were maintained using JGraphT library~\cite{Dex2007} and its APIs were used to find the shortest path (based on A*~\cite{Nilsson1968}) in the road network. The test cases were run on Intel Xeon (4 Core) processor @ 3.26 GHz with 12 GB RAM. In the following sections, the results plotted and tabulated for all the analysis have been computed as the average-value of ten iterations on the same test data.

    \subsection{Performance Evaluation on agent, plan and network characteristics}
    \indent For performance evaluation, we have used the algorithms for the Convoy Movement Problem (CMP) which are based on spatially extended objects. The performance of the optimised version of XCBS-A was compared with XCBS~\cite{Thomas2015}, the A* based solution for CMP~\cite{Thomas2008} and PDDL solution~\cite{Khemani2012}. A distributed priority based allocation strategy, along lines similar to XCBS was also compared with. Each SEA generates an optimal schedule for itself; a central conflict resolver, then does a priority ordering of the SEAs. The conflict resolution is priority based, wherein the higher priority SEA continues unhampered on the conflict area, while the lower priority SEA either waits or takes the nearest diversion. The algorithms were tested for performance to assess the impact of agent characteristic, plan characteristic and the road network characteristics. The first is captured in terms of the number of agents in the plan, the plan characteristics is evaluated in terms of the max plan length, while the road characteristics is studied in terms of road density and network size.\\

\begin{figure}[h]
    \centering
     \begin{minipage}{0.5\textwidth}
     \centering
         \includegraphics[width=1\textwidth, height=1\textwidth]{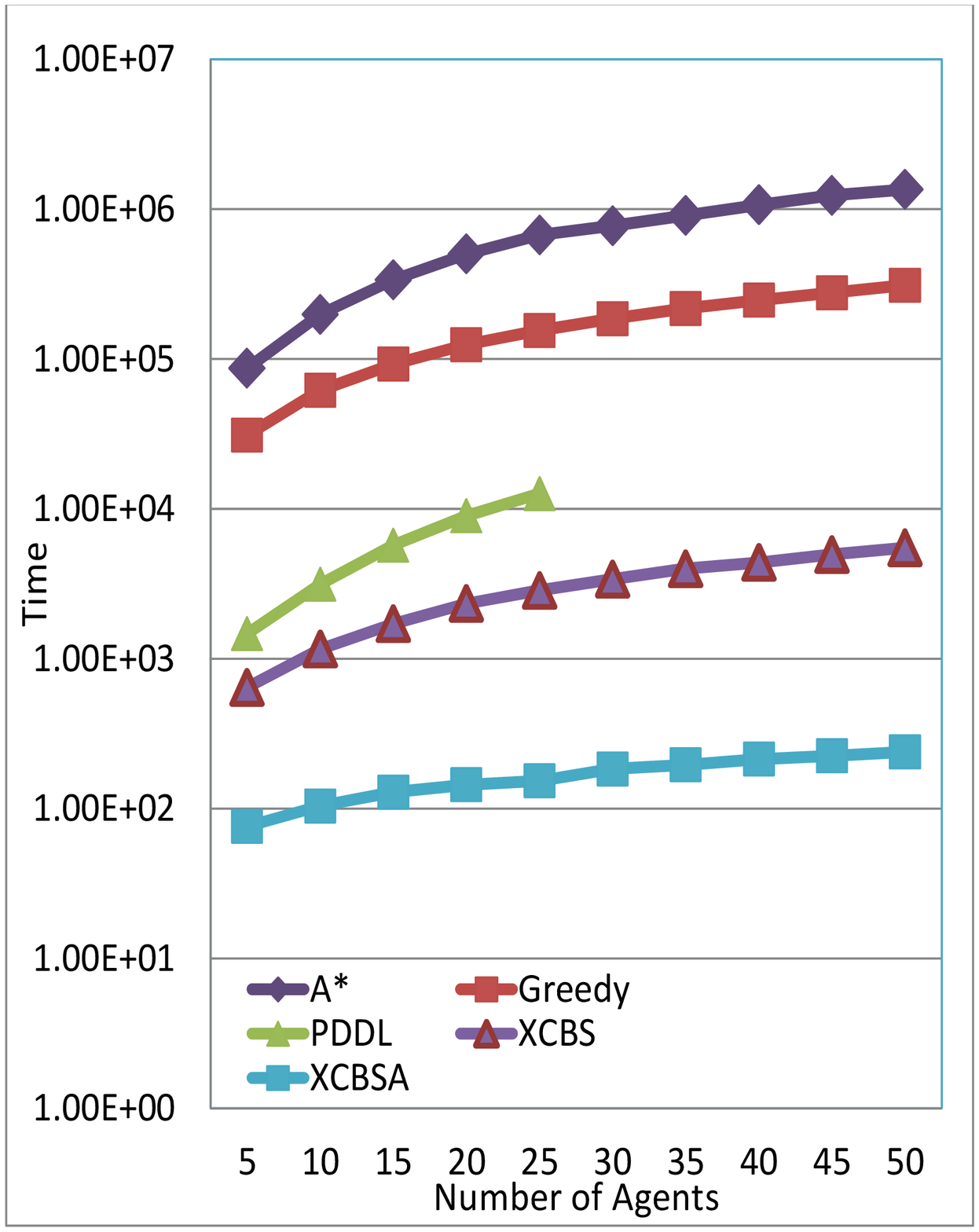}
	    \subcaption{Number of agents}
	    \label{fig:XCBSA-Agents}
    \end{minipage}%
    \begin{minipage}{0.5\textwidth}
    \centering
	    \includegraphics[width=1\textwidth, height=1\textwidth]{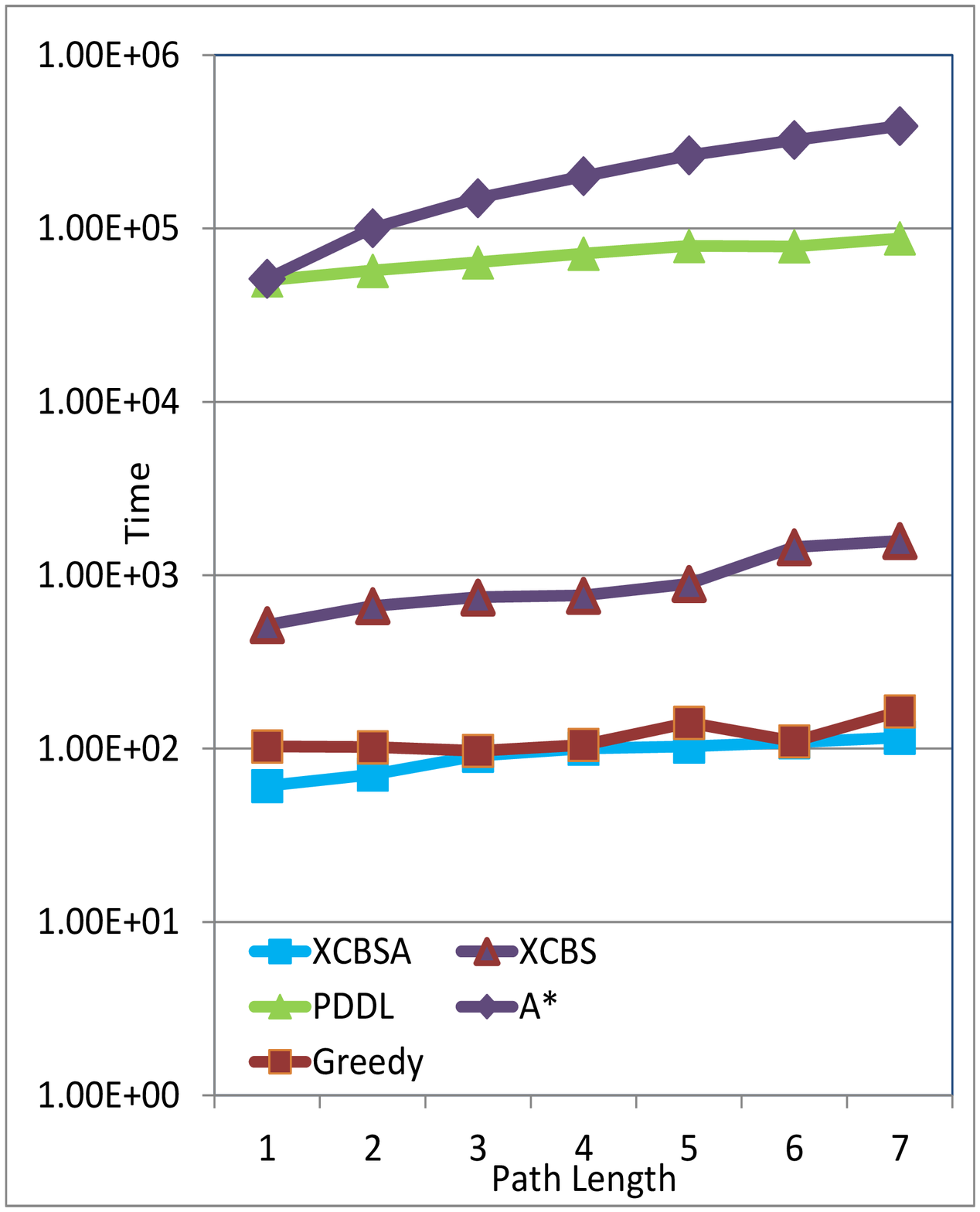}
	    \subcaption{Path Length}
	    \label{fig:XCBSA-PathLength}
    \end{minipage}
    
    \begin{minipage}{0.5\textwidth}
    \centering
	    \includegraphics[width=1\textwidth,  height=1\textwidth]{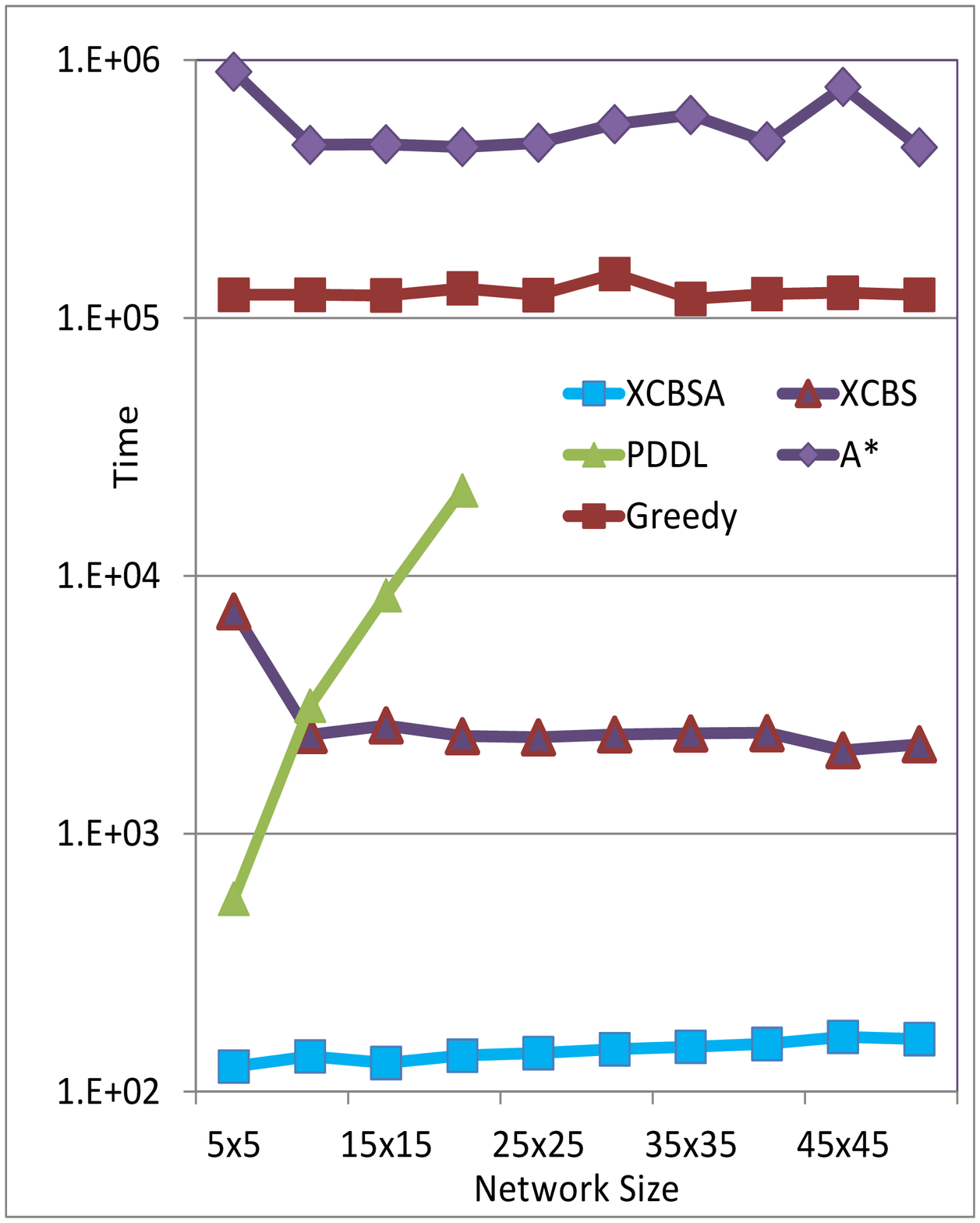}
	    \subcaption{Network Size}
	    \label{fig:XCBSA-NetworkSize}
    \end{minipage}%
    \begin{minipage}{.5\textwidth}
    \centering
	    \includegraphics[width=1\textwidth,  height=1\textwidth]{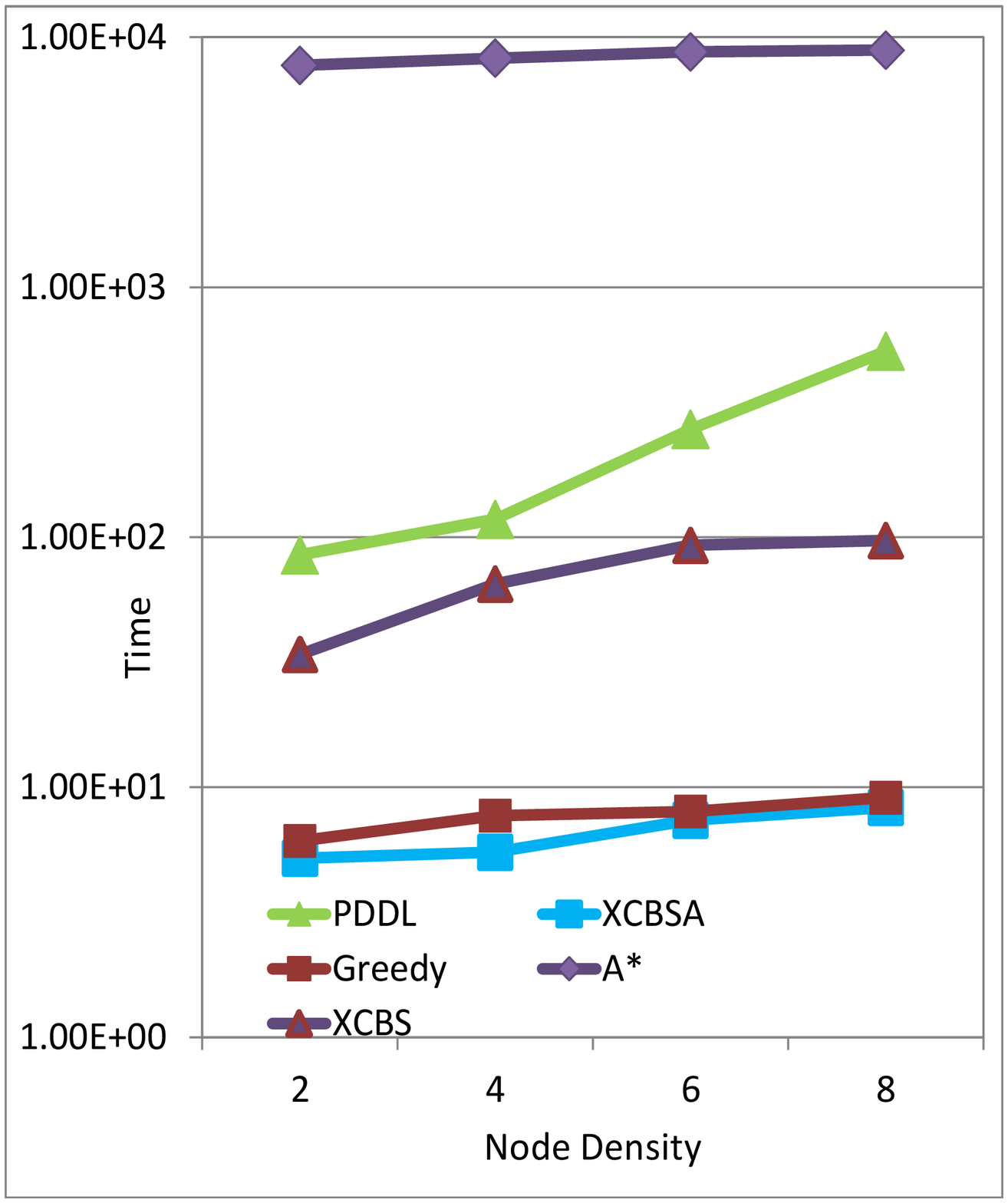}
	    \subcaption{Network Density}
	    \label{fig:XCBSA-Density}
    \end{minipage}
	    \caption{Comparative Performance of XCBS-A with other approaches}
	    \label{fig:XCBSA}

    \end{figure}

    \indent Figure~\ref{fig:XCBSA} shows the results of the executed test cases. The tests were executed on a grid based network, with uniform length roads. The agents themselves were homogeneous in terms of speed and length, however their source and destination locations were randomly decided to ensure that the path-length are uniformly maintained. In some of the tests, PDDL based solution could not be executed for all range of values as the planner used, LPG-td~\cite{LPG2004} did not support the number of predicates which were getting generated for the testcase. \\

    \indent In Figures~\ref{fig:XCBSA-Agents} to~\ref{fig:XCBSA-Density}, the comparative performance of the five algorithms indicates  XCBS-A  shows significantly better performance with respect to all the other approaches being compared with, in terms of variation in number of agents, path lengths, network sizes and network density. These results also imply that the algorithm is able to scale to solve large problem sizes without issues in terms of memory or computation time.\\

  \begin{figure}
  \begin{center}
  \begin{minipage}[t]{0.45\textwidth}
  \vspace{0pt}
  \centering
	  \includegraphics[scale=0.3]{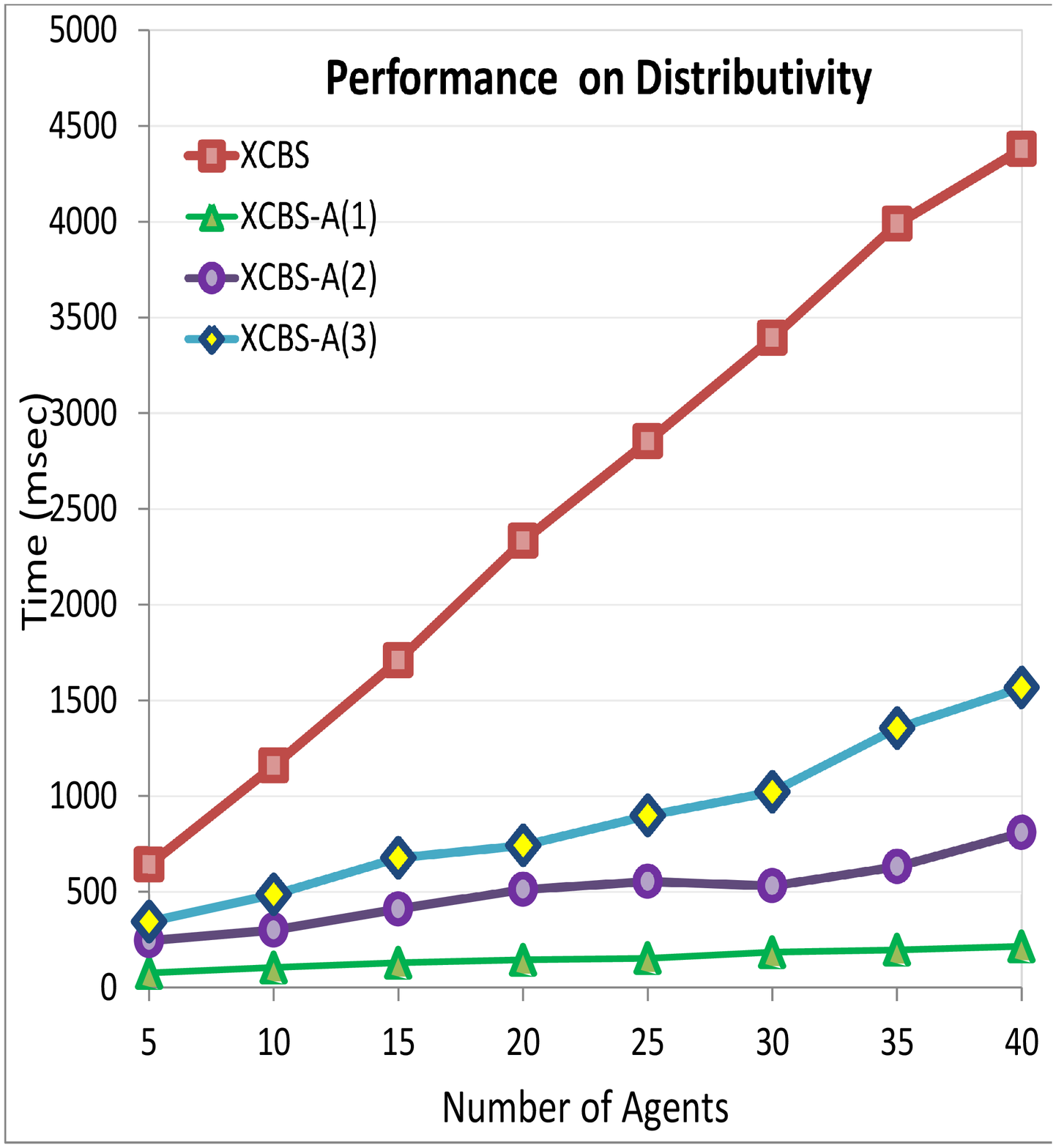}
    \caption{Evaluation of distributivity of XCBS-A}
  \label{fig:XCBSA-Distributive}
  \end{minipage}%
   \begin{minipage}[t]{0.45\textwidth}
  \vspace{0pt}
	  \captionsetup{type=table}
	  \begin{tabular}[c]{|c|c|r|r|r||r|r|r| }
		  \hline
		  \multicolumn{1}{|c|}{\multirow{2}{*}{\textbf{$\lvert A \rvert$}}}
		  &\multicolumn{1}{|c|}{\multirow{2}{*}{\textbf{$\lvert B \rvert$}}}&\multicolumn {3}{|c|}{\textbf{XCBS}} &\multicolumn{3}{|c|}{\textbf{XCBS-A}}\\
		  \cline{3-8}
		  & &\textbf{$t$} &\textbf{$N_g$} &\textbf{$N_e$} &\textbf{$t $} &\textbf{$N_g$} &\textbf{$N_e$}\\
		  \hline
		  5&1 &40757 &41&11 &\textbf{245}&\textbf{15}&\textbf{5}\\
		  7&1 &85147&85 &22 &\textbf{300}&\textbf{21}&\textbf{6}\\
		  9&1 &91565&17&27 &\textbf{554}&\textbf{36}&\textbf{8}\\
		  14 &7 &30337 &29  &8  &\textbf{490} &\textbf{129} &\textbf{2}\\
		  15 &5 &62493 &61  &16 &\textbf{649} &\textbf{179} &\textbf{3}\\
		  16 &4 &86632  &85  &22 &\textbf{708} &\textbf{164} &\textbf{4}\\
		  15 &3 &123041 &121  &31 &\textbf{785} &\textbf{171} &\textbf{5}\\
		  12 &2 &122442 &111 &31 &\textbf{602} &\textbf{71} &\textbf{6}\\   
		  14 &2 &126414 &125 &32 &\textbf{633} &\textbf{66} &\textbf{7}\\
		  16 &2 &138487 &137 &35 &\textbf{666} &\textbf{59} &\textbf{8}\\
		  \hline
	  \end{tabular}
    \caption{Comparison of Nodes generate $N_g$, nodes evaluated $N_e$  and time taken $t$}
  \label{tab:XCBSA-XCBS-Comparison}
  
    \end{minipage}
    \end{center}
  \end{figure}
    \subsection{Performance Evaluation for distributivity}
    \indent The XCBS-A is a completely distributed but centralised algorithm. For testing the distributivity and scalability, the algorithm was also implemented as a multi-agent system using Java Agent DEvelopment (JADE)~\cite{Jade2005}. JADE is a popular open source agent development framework which provides support for agent creation as well as agent monitoring, maintenance and communication. Each of the SEAs and the Central Agent was instantiated as  JADE Agents on different physical machines. The SEAs were instantiated with their initial and final locations defined, while the Central Agent was instantiated with an awareness of the total number of SEAs to participate in the planning. The communication between the SEAs and the CA was through Agent Communication Language (ACL)~\cite{fipaacl2002} message, which is the standard format for communication among agents in JADE. \\
    
    \indent The performance of XCBS-A as a multi-threaded application and as JADE based distributed deployment on multiple physical systems was compared with XCBS deployed on a single system, and is shown in Figure~\ref{fig:XCBSA-Distributive}. In the figure, XCBS-A(1) denotes multi-threaded implementation of the algorithm on a single machine. XCBS-A(2) refers to JADE implementation of the algorithm with all agents deployed on the same physical system, using unshared memory. XCBS-A(3) refers to the JADE implementation with agents distributed across three identical machines. While all three instances of the XCBS-A algorithm performed better than XCBS, the multithreaded implementation of XCBS-A was better than XCBS-A with JADE. This can be explained as the additional time taken in terms of communication between agents. This demonstrated that the algorithm is completely distributed in nature and can be scaled over multiple machines in a networked environment, seamlessly and continues to outperform XCBS.\\

  \subsection{Performance Evaluation for search space generation and exploration}
  \indent The performance of XCBS-A with respect to XCBS was evaluated in terms of time taken to arrive at the solution and the size of the search space, primarily in terms of casacading conflicts. Table ~\ref{tab:XCBSA-XCBS-Comparison} compares the performance of the two algorithms, for different number of agents $\lvert A \rvert$ and different number of blocks $\lvert B \rvert$, in terms of time taken $t$, number of CT-nodes generated $N_g$ and number of CT-nodes evaluated $N_e$.  As seen from the table, XCBS-A shows significant improvement in terms of time taken to arrive at a valid solution. Secondly, in some cases, the number of CT-nodes generated as a consequence of conflict resolution is found to be large as compared to XCBS, leading to a very huge search space, as seen in the table. There are four child CT-nodes generated for each CT-node being explored in case of XCBS, whereas in case of XCBS-A, if the partition for the CT-Node $N_e$ has $j$ non-singleton blocks with $i$ agents each, then the number of child CT-nodes generated will be $i^j$. This explains the large search tree in case of XCBS-A. However, in spite of the large tree, XCBS-A handles multiple conflicts simultaneously hence converges to a solution faster than XCBS. 
  %%%%%%%%%%%%%%%%%%%%%%%%%%%%%%%%%%%%%%%%%%%%%%%%%%%%%%%

  \section{Memory Efficient XCBS-A}\label{xcbsa-eff}
    \indent In XCBS-A, at each level of the CT there are $i^j$ CT-nodes which may be of the same cost. As the cost of the CT-nodes is non-decreasing down the tree, all CT-nodes of same cost at a level of the CT would have to be explored before higher-cost CT-nodes, in lower branches, could be explored. This leads to the generation and exploration of the large search space as was seen in the results discussed above. We looked into anytime approaches \cite{Thayer2010} which try to arrive at optimised solutions by iteratively expanding the search space based on the deliberation time available. In these approaches \cite{Cohen2018,Pearl1982,Thrun2003} different sets were used to maintain the explored, unexplored, inconsistent and select nodes of the search space.\\
    
    \indent  Inspired by the above mentioned approaches, we formulated a strategy of adding only potentially explorable CT-nodes into the Open set. Meta-data for regeneration of other child CT-nodes is stored in another set called Potential Set (PS). The revised high-level search is as shown in Algorithm \ref{algo-memory-eff}. As per the strategy, when a CT-node $N_e$ is being chosen to be explored, it is compared with the best-node from the Potential Set. If cost of the CT-node from Potential Set is lesser than the cost of the node from the Open Set, then it is removed from Potential Set and explored. This is shown in Line 6-10 of the algorithm.  Further, on conflict resolution from among the multiple child CT-nodes generated, only one least cost node is added to the Open Set. The metadata with respect to the remaining nodes is placed in the Potential Set. This approach ensures that the CT comprises only one of the least-cost CT-nodes generated by the exploration of a CT-node. Hence the search space is  reduced significantly. Moreover since only meta-data of a child node is maintained in Potential Set, hence it does not consume much storage space.\\
  \begin{algorithm}
	    \caption{\textbf{Memory-Efficient High Level Search}}\label{algo-memory-eff}
	    \begin{algorithmic}[1]
		    \State Define root CT-Node $N_r$ with optimal consistent plan for each $a_i$ agents.
		    \State OpenSet $O \gets \{N_r\}$ 
		    \State Potential Set $PS \gets \phi$
		    \State $ResultFound \gets false$.  
		    \While {!$O.empty$ or !$ResultFound$} 
		    \If{$min_{N^{'} \in O} N^{'}.Cost <= min_{\tilde{N} \in PS} \tilde{N}.Cost}$
		    \State $N_e \gets arg min_{N^{'} \in O} N^{'}.Cost$
		    \State $O \gets O \setminus {N^{'}}$
		    \Else
		    \State $N_e \gets PS.getBestNode()$
		    \EndIf
		    \State $conflictPresent \gets Validate(N_e)$
		    \If {$conflictPresent$}
		    \State $\{ChildCTNodes\} \gets ResolveConflict(N_e)$
		    \State $N_{min} \gets arg min_{N^{'} \in ChildCTNodes} N^{'}.Cost$
		    \State $O \gets O \cup \{N_{min}\}$
		    \State $PS \gets ChildCTNodes\setminus N_{min}$
		    \Else 
		    \State $ResultFound \gets true$
		    \EndIf
		    \EndWhile  
		    \If {$ResultFound$}
		    \State $Solution \gets N_e.Solution$
		    \EndIf
	    \end{algorithmic}
    \end{algorithm}

    \begin{figure}
      \includegraphics[width=\textwidth]{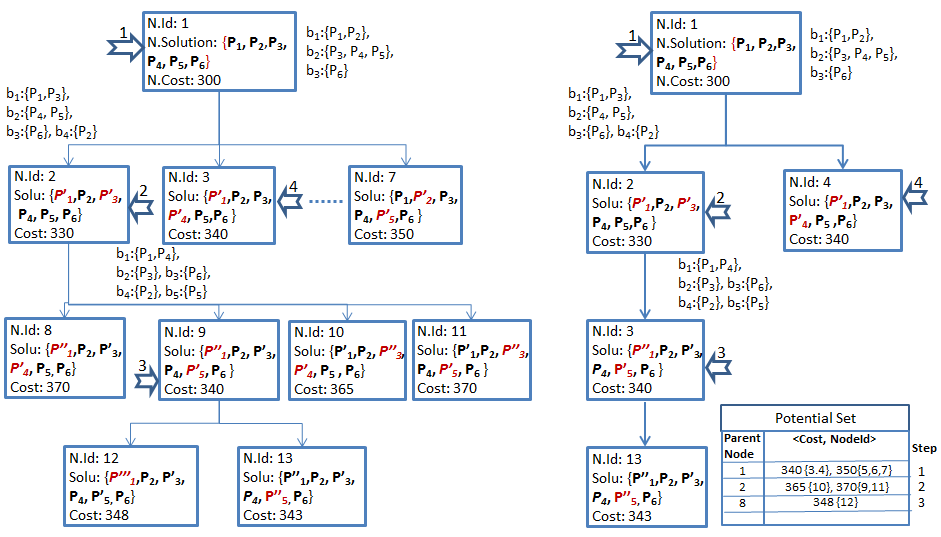}
    \caption{Figure shows comparative generation and exploration of CT in XCBS-A (left) and XCB-A-Eff (right).}
    \label{fig:xcbsaeff}
      \end{figure}

    \indent \textbf{Example:} Figure \ref{fig:xcbsaeff} shows a small illustrative example indicating the growth of the CT in the normal mode and in the optimised mode. The root CT-Node, $N_r$, has the partition as shown in the figure. In Step 1 (indicated by arrow numbered 1), CT-node.Id 1, is explored to generate the CT-nodes with combined revised plans of agents from blocks $b_1$ and $b_2$ of the partition.  In normal CT (shown in the left figure), this leads to generation of CT-nodes 2 to 7. In the optimised mode (shown in the right figure) only one CT-node, which is the least cost node among all solutions is added to the CT. The remaining solutions are maintained cost-wise at the Potential Set. Thus as a consequence of Step 1, in optimised mode, the Potential Set makes entries with costs 340 and 350, for the parent CT-node 1. The CT-nodes explored $2^{nd}$ and $3^{rd}$ are indicated by the corresponding numbered arrowheads. Consequent to Step 3, the least cost node in the optimised mode CT is of cost 343, while the least cost node available in the Potential Set is of cost 340. This node is then extracted from the Potential Set and added to the optimised CT, with its parent as defined in the Potential Set. The node (marked with arrowhead 4) is then explored, and the entire search process is consistent across both modes of operations. The reduced search space is evident from the example.\\
    
    \indent The Potential Set ends up having several CT-nodes with the same cost and hence choosing one from among them as the best-node is guided by a heuristic measure. We attempted three different heuristic measures for the same. 
    \begin{itemize}
	    \item Deeper-first: The child node which is to be positioned deeper, would have resolved more conflicts as compared to other nodes of the same cost, on higher branches. This would favour search paths of the tree where search has progressed more.
	    \item Largest-block first: The child node, the partition set of whose parent had the largest block. The idea being that resolving the group with the largest mutually conflicting agents will likely lead to solution faster.
	    \item Most-singletons first: The child node, the partition set of whose parent had the maximum number of singletons. The idea being that greater the number of singletons, the closer the node is to the solution.
    \end{itemize}
    However, preliminary experimentation revealed that the XCBSA-Eff approach gave an advantage in terms of the total number of nodes in the CT. Further, in terms of the number of nodes explored to arrive at the solution, Deeper-first heuristic gave better performance as compared to the other heuristics. Hence for the rest of the paper, all our references to XCBSA-Eff will be with respect to the Deeper-first heuristic. \\
    
  \begin{table}
  \resizebox{\textwidth}{!}
  {
      \begin{tabular}[c]{|c||c||r|r|r||r|r|r||r|r|r|| }
		    \hline
		    \multicolumn{1}{|c||}{\textbf{$\lvert A \rvert$}}
		    &\multicolumn{1}{|c||}{\textbf{$\lvert B \rvert$}}&\multicolumn {3}{|c||}{\textbf{Nodes Generated}} &\multicolumn{3}{|c||}{\textbf{Nodes Evaluated}}&\multicolumn{3}{|c||}{\textbf{Time Taken}}\\
		    \cline{3-11}
		    & &{XCBS} &{XCBSA} &{XA-Eff} &{XCBS} &{XCBSA} &{XA-Eff} &{XCBS} &{XCBSA} &{XA-Eff}\\
		    \cline{1-11}
		    \multirow{3}{*}{10} &2 &41 &32 &7&23 &19 &7& 1011 &368 &248\\
		    &3 &56 &39 &9 &33 &21 &9& 1345 &381 &328\\
		    &5 &81 &55 &6 &46 &38 &6& 351 &352 &285\\
		    \cline{1-11}
		    
		    \multirow{3}{*}{20} &2 &125 & 108 &9&32 & 20&9& 1823 & 1048   &906\\
		    &3 & 156 & 132 &9 &76 &48 &7& 2047 &1429 &1081\\
		    &5 &161 &159 &6 &41 &35 &6& 1893 &1251 &843\\
		    &10 &218 &189 &15&125  &78  &10 &4924  &1783 &1294\\
		    \cline{1-11}
		    
		    \multirow{3}{*}{30} &2 &173 &152&18&44 & 27&18&7591 &3561 &2680\\
		    &3 &212 &173 &15 &119 &84 &12 &9334 &5839 &3901\\
		    &5 &225 &169 & 23&157 &125 &17&12590 &7598 &5998\\
		    &10 &318 &228 & 43&216 &163 &37&23459  &10231 &8921\\
		    &15 &572 &432 & 71&347 &216 &56&59011  &32054 &12116\\
		    
		    \hline
		  
	    \end{tabular}
	    }
	    \caption{Evaluation of XCBS, XCBS-A , XCBSA-Eff with Deeper-first heuristic (XA-Eff) for CT-nodes generated, CT-Nodes explored and time taken.}
	    \label{tab:XCBSA-Eff Eval}
    \end{table}

  \subsection{Results}
  The comparative performance of the efficiency of the new approach XCBSA-Eff, with the Deeper-first heuristic, was carried out with respect to XCBS and XCBS-A. The primary difference expected was in terms of the number of CT-nodes generated and number of CT-nodes evaluated. The test suite comprised a grid based network, with uniform length roads. The agents themselves were homogeneous in terms of speed and length, however their source and destination locations were randomly decided to ensure that the path-length are uniformly maintained. \\

  \indent The Table~\ref{tab:XCBSA-Eff Eval} captures the comparative performance of the three approaches. It is seen that overall, XCBSA-Eff performs better than XCBS and XCBSA, in terms of the number of nodes generated, evaluated and consequently time taken to arrive at the solution. However we bring out a few points to note. First, the difference between the number of nodes generated and evaluated in XCBSA-Eff is not very significant. This is primarily because the homogenity of the road network and agent characteristics leads to large number of alternative solutions of the same cost, which are all put in the Potential Set. In a non-uniform road network, the number of nodes generated and evaluated in terms of XCBSA-Eff is expected to be greatly different, however it will always be lesser than that generated by either XCBS or XCBS-A. The second point is that  the number of cascading conflicts has a direct influence on the number of nodes which are generated and explored. This is seen in some rows, where though the number of agents and blocks were higher, lesser number of cascading conflicts resulted in lesser number of nodes being generated and explored both for XCBSA and XCBSA-Eff.

  %%%%%%%%%%%%%%%%%%%%%%%%%%%%%%%%%%%%%%%%%%%%%%%%%%%%%%%
	    
    \section{XCBS-Local Awareness (XCBS-LA)}\label{xcbsla}
    In XCBS-A, one agent from every conflict block is replanned while the plans of all other agents remain unchanged. This ensures that the search down a branch terminates but the drawback is that singleton agents' plans remain unchanged, which generates a bias in favor of the singleton agent blocks. This bias sometimes keeps the algorithm from arriving at an optimal solution. \\
  
    \indent Towards this, we propose a revised approach, XCBS-Local Awareness (XCBS-LA), which explores all possibilities of plans without a bias and  leads to an optimal solution. The principle difference between XCBS-LA and XCBS-A  lies in conflict resolution, which happens among agents within the same partition block, without awareness of agents of other blocks. Since in this mechanism any agent is looking to resolve conflicts only with agents in its own block, we have termed this approach as XCBS-Local Awareness. The conflict resolution leads agents in the same block to make a \textit{commitment} to the resolved plans of each other, which is maintained down the branch of the Constraint Tree.\\
    \begin{definition}
    A \textit{commitment} $C$, is a binary relation on $A$, such that
    \begin{equation}
    aCb = \{constraints(b) \cup plan(b)| a,b \in A\}
    \end{equation}
    
    The commitment relation implies that the plan of agent $a$ satisfies all constraints of agent $b$ and those imposed by the plan of agent $b$. By definition itself, $C$ is asymmetric, and as it maintains the commitments it made with itself in the parent CTnode, hence $C$ is also reflexive.
    
    \end{definition}

  \indent In XCBS-LA, the High-Level Search and Validate procedures remain the same as in XCBS-A, however, the Conflict Resolution is revised as discussed in the following section. 

  \begin{algorithm}
    \caption{\textbf{ResolveConflict $N_e$}}\label{resolveConflict-LA}
    \begin{algorithmic}[1]
	    \Procedure{ResolveConflict}{$N_e$}
	    \For {each non-singleton block $b_i$ in partition} 
	    \State $P^{'}_{i}$ = BlockLevelSearch($b_i, N_e$).
	    \EndFor 
	    \State $\mathfrak{B} \gets $combinations of $P^{'}_{i}$
	    \For {each $\mathfrak{b} \in \mathfrak{B}$ } 
	    \State  Child CT-node $N_c  \gets P^{'}_{b}$.
	    \State Compute Cost($N_c$).
	    \State $O \gets O \cup N_c$
	    \EndFor 
	    \EndProcedure  
    \end{algorithmic}
    \end{algorithm}
    \subsection{Conflict Resolution}
    \indent Once a partition of the agents into blocks of mutually conflicting agents is created, in XCBS-LA, the ResolveConflict is invoked as shown in Algorithm~\ref{resolveConflict-LA}. In XCBS-LA, the conflicts are resolved by ensuring that agents within the same block generate an optimal conflict free plans with respect to each other, ignoring the plans of agents in other blocks. Any optimal MAPF algorithm can be adopted for generating plans for the agents in the same block. We have used XCBS in our implementation. Thus the BlockLevelSearch in Line 3 of the algorithm, executes XCBS on the agents in a block. The revised plans $P^{'}_{i}$  are composed by the Central Agent into child CT-node. If $N_e$ resulted in $j$ non-singleton blocks of partition, there will be $j^2 -1$ child CT-nodes generated.\\

    \subsection{Block Level Search}
    \indent An agent $a_i$ makes a \textit{commitment} with an agent $a_j$, if the plan and constraints of $a_j$ are satisfied by $a_i$ to generate a consistent solution. In Block Level Search, the XCBS algorithm is invoked for each non-singleton block of the partition. The XCBS call returns an optimal conflict-free solution for all the agents in the block. This results in each agent establishing a commitment with respect to the other agents and their generated plan. Once a commitment is established, thereafter, anytime an agent $a_i$ is replanned by XCBS, it will get its constraints with respect to the current constraints and plans of all other agents $a_j$ with which it has a commitment.\\
    
    \subsection{Theoretical Analysis}
    \begin{lemma} \label{XCBSLA-Lemma1}Given any two agents $a_i$, $a_j$, if $a_i$ makes a commitment with respect to $a_j$ in a branch of the CT, it will not conflict with it further down in that branch. \end{lemma}
    
    \begin{lemma} \label{XCBSLA-Lemma2}Down the branch of a tree, the commitments of an agent with others, leads to more constraints in its plans and hence the cost of the individual plan will be non-decreasing down a branch of the CT.\end{lemma}
	    
    \begin{lemma}If $\mathfrak{S}$ be the set of all optimal solutions for a given MAPF instance, then any $S \in \mathfrak{S} $ will be generated by XCBS-LA in the CT.\end{lemma}
    Let us consider the case by contradiction, let $\exists S \in \mathfrak{S}$, such that $S$ is not in CT.  Since $S$ comprises conflict free plans of all the agents, it implies that the plan of atleast one of the agents, say $a_i$, is different from the solutions in the CT. If $P(a_i)$ is its most optimal plan, then it will be there in $N_r$ of CT. Further, if it does not conflict with any other agent plan, then $P(a_i)$ will also be present in the child-CT nodes, because plans are changed only if there is a conflict. Hence if $P(a_i)$ is the optimal plan of the agent $a_i$ it will be in CT. Now we consider the second case, i.e. $P(a_i)$ is not the most optimal plan of agent $a_i$, which implies it has conflicts with one or more agents, say $a_j$. Now XCBS generates the optimal plan for the block of agents comprising $a_i$ and $a_j$ and gives the optimal plan for the block. Since XCBS is proven to be optimal, the plan generated by XCBS for $a_i$ will be optimal and will be part of CT. Thus, in both the cases, the assumption is contradicted.\\

    \textbf{Theorem 1} \textit{XCBS-LA is optimal. }\\
    Proof: The CT contains all the possible combinations of plans of the agents as CT-nodes, which are explored based on their increasing cost. When a least code CT-node is found to be valid, all other remaining CT-nodes to be explored, are either of the same cost or higher. Therefore any other valid unexplored solution will either be of the same cost or higher as per Lemma~\ref{XCBSLA-Lemma2}. Hence the solution found is optimal. \\

    \textbf{Theorem 2} \textit{XCBS-LA is complete. }\\
  Proof: A CT-node can expand into a finite number of child-CT nodes. Further, on exploring down a branch of a tree, as  per Lemma~\ref{XCBSLA-Lemma1}, once an agent makes a commitment, it will not conflict with the other agent down the tree. Hence either all agents eventually will form commitments with each other, or remain mutually exclusive, thereby resulting in a valid solution.
    
    \subsection{Results}
  The performance of XCBS-LA was evaluated against four approaches namely the priority based Greedy algorithm, the XCBS, the XCBS-A and XCBS-A(Eff). A test-suite for a 100$\times$100 grid network with 50 agents was generated such that in their initial plans the agents could be split into blocks of definite sizes. These block sizes were a key factor in the comparison of the performance of the algorithms.

\begin{figure}
    \begin{subfigure}{0.4\textwidth}
	    \includegraphics[scale=0.33]{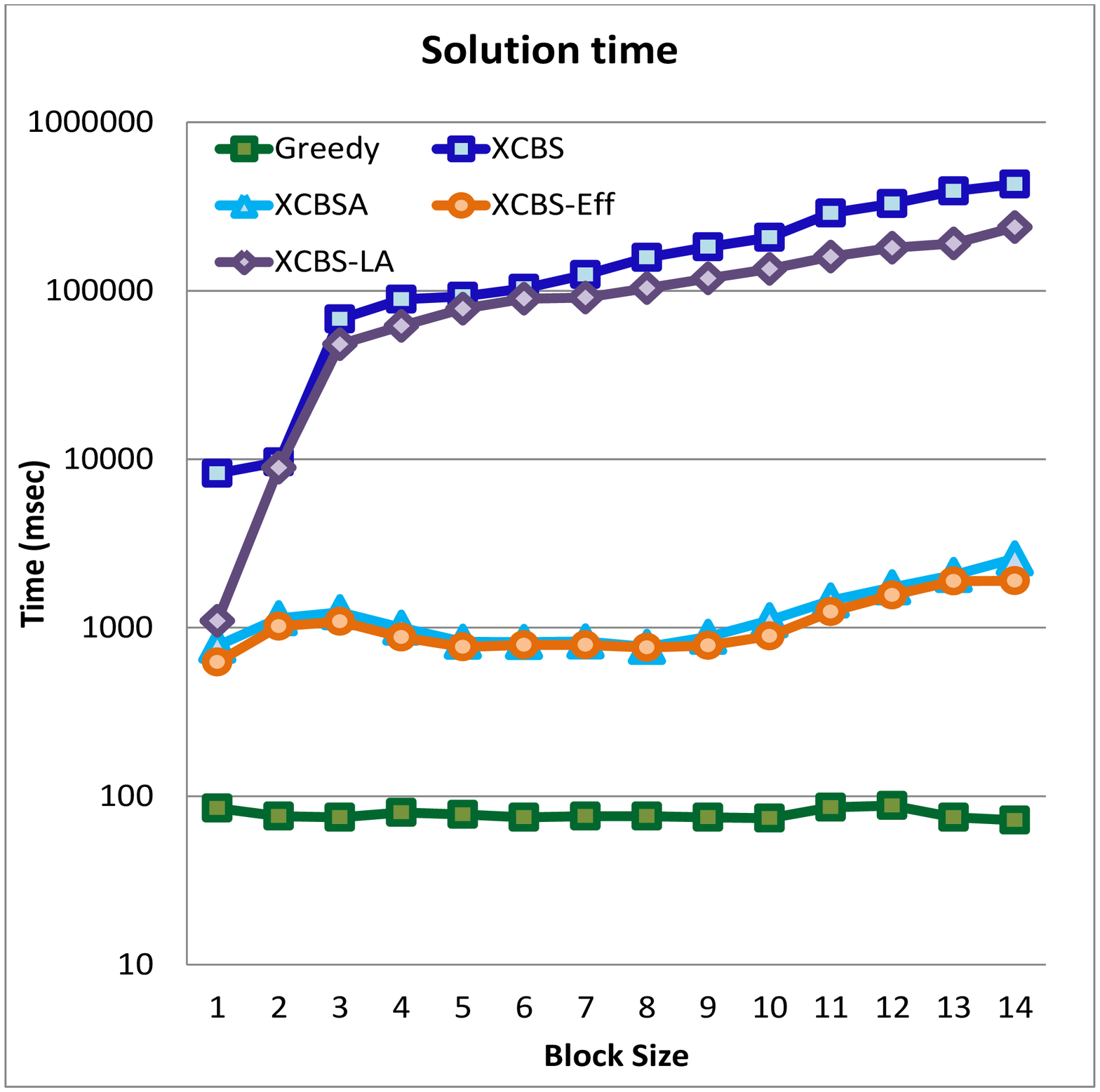}
	  
	    \caption{Comparison of time taken to arrive at solution}
	    \label{fig:XCBSLA_time}
    \end{subfigure}
  \hspace{35pt}
    \begin{subfigure}{0.4\textwidth}
	    \includegraphics[scale=0.33]{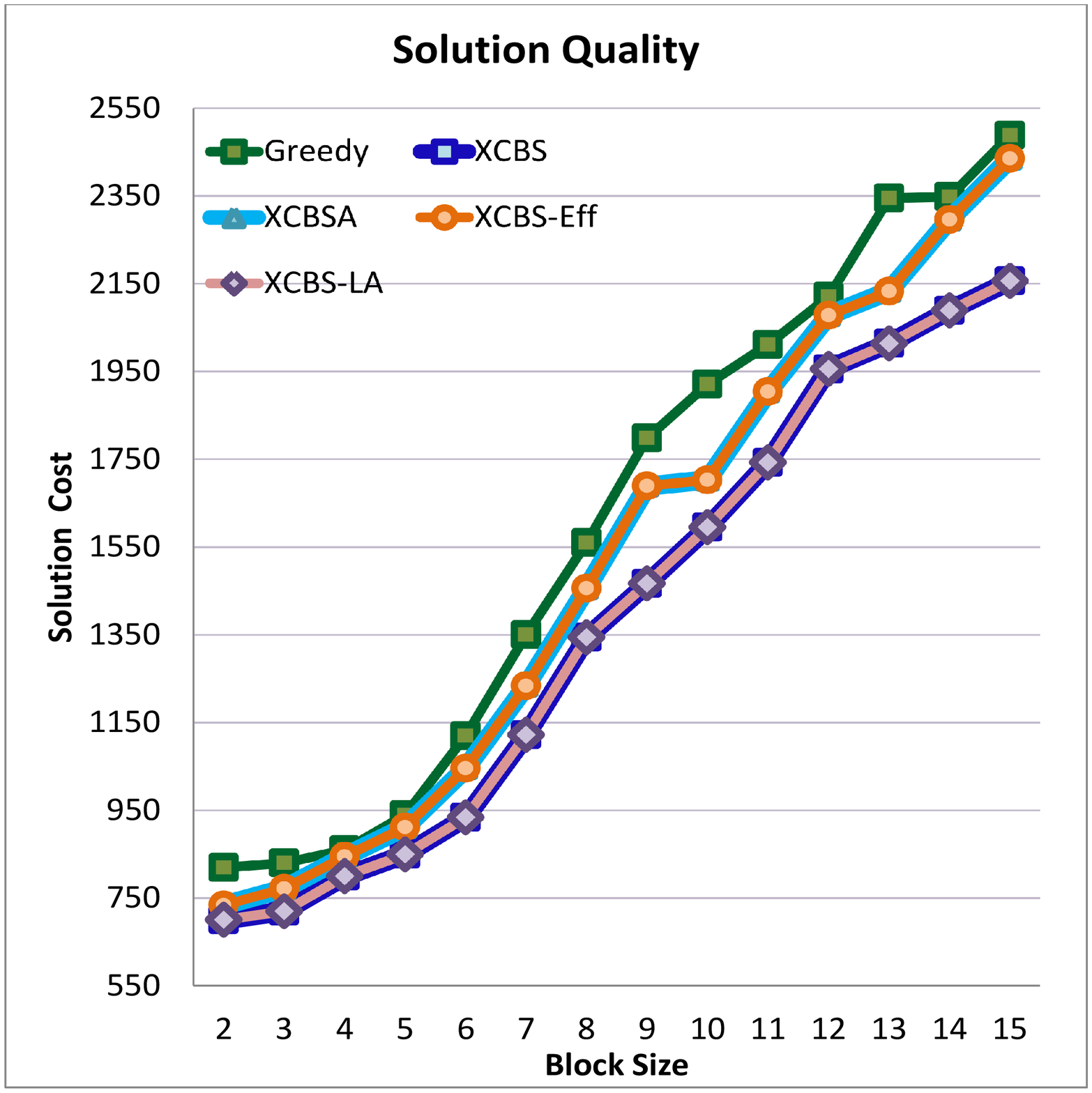}
	  
	    \caption{Comparison of solution quality, in terms of Plan Cost. }
	    \label{fig:XCBSLA_qlty}
    \end{subfigure}%

	    \caption{Figure shows the comparative performance of XCBS-LA with respect to other approaches. }
	    \label{fig:XCBS_LA}
 \end{figure}

  \indent The evaluation results are captured in Figure~\ref{fig:XCBS_LA} for the time taken and the quality of the solution. As seen in the Figure~\ref{fig:XCBSLA_time}, in terms of time taken to arrive at a solution, XCBS-LA performs poorly as compared to XCBS-A or XCBS-A(Eff), though better than XCBS. This can be attributed to the fact that it internally invokes XCBS, which takes time depending on the number of agents in the block. The quality of the solution is compared in terms of the sum-of-plan costs. It is observed that XCBS-LA gives the optimal solution, same as XCBS and better than the other approaches, as shown in Figure~\ref{fig:XCBSLA_qlty}. 

  %%%%%%%%%%%%%%%%%%%%%%%%%%%%%%%%%%%%%%%%%%%%%%%%%%%%%%%

    \section{CONCLUSION}\label{conclusion}
    In this paper, we have looked into an approach for multi agent path finding for spatially extended agents. We have analysed a distributed algorithm called eXtended Conflict Based Search with Awareness(XCBS-A) for conflict-free path finding of spatially extended agents. XCBS-A was successfully tested for several agents over different types of networks. The comparative performance of XCBS-A with four other approaches in terms of agent characteristics, plan characteristics and road network characteristics indicate that XCBS-A performs significantly better than the other algorithms. We have been able to demonstrate the performance of the algorithm over distributed machines using a multi agent framework JADE, which makes it a good solution for on-field deployment. \\
     
    The memory footprint of the XCBS-A algorithm was reduced by a memory-efficient version of the XCBS-A which tries to prune the search tree, CT, by pushing nodes of similar costs into a Potential Set. This reduces the size of the CT and also helps in reducing the number of nodes evaluated to arrive at a solution. \\	
     
    Finally, we proposed another approach namely, the  eXtended Conflict Based Search with Local Awareness (XCBS-LA). This algorithm overcomes the bias which XCBS-A introduces in favour of singleton agents or agents which do not conflict with any other agents. This algorithm guarantees optimal solution though at a cost of a higher processing time as compared to XCBS-A. \\
    
    In further steps, we need to look into using the network and the agent characteristics as heuristics to prune the CT in case of XCBS-LA and lead to reduction in the processing time. Further, the performance of these algorithms varies depending on the number of casacading conflicts, nature of the road network etc. Hence a framework which could analyse the characteristics of the problem and then choose the appropriate algorithm for an efficient and quick solution is another area to be looked into. Another direction of research also needs to be in the line of faster conflict-detection and resolution. These topics remain the focus of future research work in this area.
    
  %%%%%%%%%%%%%%%%%%%%%%%%%%%%%%%%%%%%%%%%%%%%%%%%%%%%%%%

  \bibliographystyle{apacite}
  \bibliography{xswa-xcbsa}

  %%%%%%%%%%%%%%%%%%%%%%%%%%%%%%%%%%%%%%%%%%%%%%%%%%%%%%%

  \end{document}